\documentclass[10pt,twocolumn,letterpaper]{article}
\pdfoutput=1
\usepackage{iccv}
\usepackage{times}
\usepackage{epsfig}
\usepackage{graphicx}
\usepackage{amsmath}
\usepackage{amssymb}

\usepackage{multirow}
\usepackage{algorithm}
\usepackage{algorithmic}

\usepackage[pagebackref=true,breaklinks=true,letterpaper=true,colorlinks,bookmarks=false]{hyperref}

\iccvfinalcopy 

\def\iccvPaperID{690} 

\ificcvfinal\fi
\begin{document}

\title{Hyper-Fisher Vectors for Action Recognition}

\author{Sanath Narayan  \qquad Kalpathi R. Ramakrishnan \\
Dept. of Electrical Engg., Indian Institute of Science,
Bangalore\\
{\tt\small \{sanath,krr\}@ee.iisc.ernet.in}
}

\maketitle

\begin{abstract}
   In this paper, a novel encoding scheme combining Fisher vector and 
   bag-of-words encodings has been proposed for recognizing action in videos.
   The proposed Hyper-Fisher vector encoding is sum of local Fisher vectors 
   which are computed based on the traditional Bag-of-Words (BoW) encoding. 
   Thus, the proposed encoding is simple and yet an effective representation
   over the traditional Fisher Vector encoding.
   By extensive evaluation on challenging 
   action recognition datasets, {viz}., Youtube, Olympic Sports, UCF50 and HMDB51,
   we show that the proposed Hyper-Fisher Vector encoding 
   improves the recognition performance by around $2-3\%$ 
   compared to the improved Fisher Vector encoding. 
   We also perform experiments to show that the performance of the 
   Hyper-Fisher Vector is robust to the dictionary size of the BoW encoding. 
\end{abstract}


\section{Introduction}

   Recognizing actions in videos has been an important topic of research for long.
   It is required in applications like automatic video retrieval and indexing, video surveillance, 
   suspicious activity detection, sports video analysis, personal gaming, behavior monitoring of patients \etc {}
   The various challenges in recognizing actions include variations in the environment, 
   intra-class variations, high-dimensionality of data.
   Changes in the environment include moving background (cars, pedestrians), changes in camera view-points,
   dynamic background due to moving camera, occlusion to name a few. 
   
   \begin{figure}[t]
      \begin{center}
	\includegraphics[width=0.95\columnwidth,height=1\columnwidth]{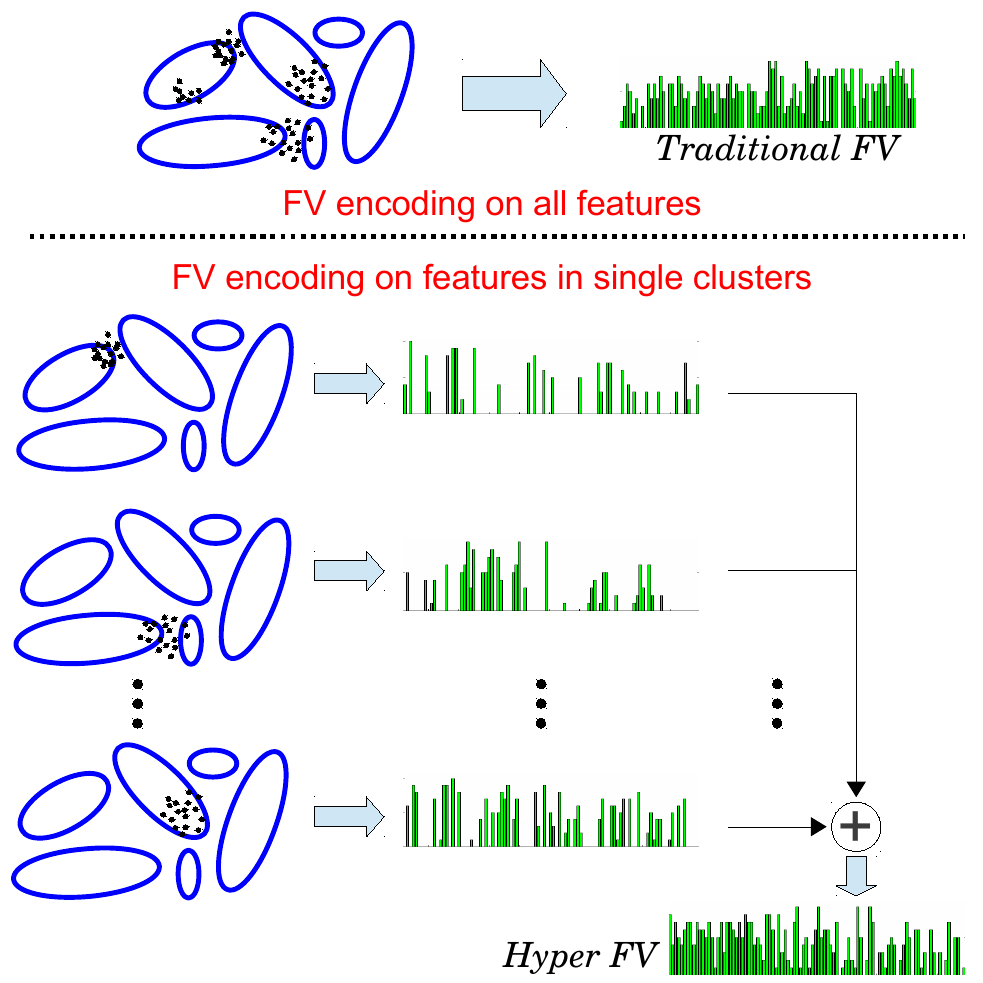}
      \end{center}
      \caption{Illustration of proposed HFV encoding in comparison to the traditional FV encoding.
               In the proposed Hyper-Fisher Vector approach, features in different clusters are 
               FV encoded separately and summed resulting in a better representation of the variations in the 
               features than the traditional FV.}
      \label{fig:proposed_hfv}
   \end{figure}	   
   
   The commonly used Bag-of-Words (BoW) representation \cite{BoW03Object} 
   consists mainly of feature extraction, generating codebook, 
   feature encoding and pooling, and normalization. 
   Development of well designed low-level features 
   like dense trajectory features \cite{Wang13ijcv,Wang13iccv} and more sophisticated 
   encoding schemes like Fisher vector encoding \cite{Fisher10} has resulted in 
   the good performance of BoW model.
   Though, Fisher vector (FV) encoding \cite{Fisher10} is also a variant of BoW model, for clarity purposes, BoW 
   represents (from here on) the Bag-of-Words encoding using \emph{k}-means clustering.
   Our proposed work uses the popular improved trajectory features \cite{Wang13iccv} and focuses on improving 
   the encoding of the features to improve the recognition performance. The proposed encoding is based on 
   embedding the BoW encoding into the FV encoding.
   The proposed encoding is simple and effective and robust to variations in 
   the dictionary size of BoW encoding.
   This modification can be used in general for other applications, 
   apart from action recognition, which use FV encoding for descriptor
   computation. The framework of the approach in comparison to the traditional FV encoding
   is illustrated in figure~\ref{fig:proposed_hfv}.


   \subsection{Related Work}
    Predominantly, there have been many methods to classify actions using low-level features 
    based on space-time interest 
    points (STIP) using various detectors based on Harris3D \cite{Laptev03}, 
    separable Gabor filters \cite{Dollar05}, etc. 
    The local features describing the interest points are
    generally based on gradient information, optical flow \cite{Dollar05, Laptev08, Schuldt04, 
    Wang09}, local trinary patterns \cite{Yeffet09}, 3D-SIFT \cite{Scovanner07}. 
    Few of the other approaches include 
    space-time shape representations \cite{Gorelick07} and template-based methods 
    \cite{Bobick01, Efros03, Rodriguez08, Corso12}. 
  
    In recent years, the trajectory-based methods to perform action classification
    have become popular and 
    are presented in \cite{Ali07, Trajectons09, Wu11, Jiang12, Wang13ijcv, Wang13iccv}.
    Ali \etal \cite{Ali07} used chaotic invariants as features on manually obtained trajectories
    to recognize actions.
    Harris3D interest points are tracked and temporal velocity 
    histories of trajectories are used as features by Messing \etal \cite{Messing09}.
    Matikainen \etal \cite{Trajectons09} used sparse trajectories from KLT tracker 
    with elements of affine matrices in bag-of-words context as features. 
    However, the performance of dense trajectories is observed to be better than sparse 
    trajectories \cite{Wu11, Wang13iccv}. Wang \etal \cite{Wang13ijcv} use local 
    3D volume descriptors based on motion boundary histograms (MBH) \cite{Dalal06},
    histogram of oriented gradients (HOG) and histogram of optical flow (HOF)
    around dense trajectories to encode action. 
    Recently in \cite{Wang13iccv}, Wang \etal estimate the camera motion 
    and compensate for it and thereby improving the trajectories and the associated descriptors.
    The interactions between the dense motion trajectories in an action are quantified and used for recognising actions 
    in \cite{Jiang12,Narayan14cvpr}.
    
    Related to our work of encoding features, Peng \etal \cite{BoWFusion14} give a comprehensive study 
    of the fusion methods for different encoding schemes for action recognition. They evaluate the 
    performance of different encodings, pooling and normalization strategies and fusion methods.
    Three kinds of fusion levels, \emph{viz}., descriptor-level, representation-level and score-level
    fusion are studied. A hybrid representation of fusing outputs from different encodings is also given.
    Of the three fusion methods, representation-level fusion is closer to our proposed work.
    The representation-level fusion and the fusion used in hybrid representation
    are outside of the encoding schemes, unlike in this work, where 
    we are incorporating one encoding (BoW) with in another encoding (FV). 

    The contribution of this paper is a
    novel and effective Fisher Vector encoding which performs 
    better than the traditional Fisher Vector encoding.
    Organization of the rest of the paper is as follows. The Hyper-Fisher Vector
    encoding for action representation is explained in Section~\ref{sec:hfv}. 
    The details of Experimental setup are provided in Section~\ref{sec:exp_setup}. 
    Results on various datasets for action recognition and experiments related to the robustness of the
    Hyper-Fisher Vector encoding
    are given in Section~\ref{sec:exp_results} and we conclude the paper in Section~\ref{sec:conclusion}.

  \section{Hyper-Fisher Vector Encoding}
  \label{sec:hfv}
    In this section, the proposed Hyper-Fisher Vector encoding is detailed. At first, 
    Fisher Vectors are explained briefly in section~\ref{sec:fv_basics}. 
  
  \subsection{Fisher Vectors}
  \label{sec:fv_basics}
    Derived from Fisher kernel, Fisher Vector (FV) coding method was originally
    proposed for large scale image categorization \cite{Fisher10}. The assumption in FV
    encoding is that the generation process of local descriptors $\mathbf{X}$ can be modeled 
    by a probability density function $p(·; \theta)$ with parameters $\theta$.
    The contribution of a parameter to the generation process
    of $\mathbf{X}$ can be described by the gradient of the log-likelihood with respect to that
    parameter. Then the video can be described by 

    \begin{equation}
      G_{\theta}^{\mathbf{X}} = \frac{1}{N} \nabla_{\theta}\log p(\mathbf{X};\theta)
    \end{equation}
 
    The probability density function is usually modeled by Gaussian Mixture
    Model (GMM), and $\theta = \{\pi_k , \mu_k, \sigma_k : k = 1 \ldots K \}$ 
    are the model parameters
    denoting the mixture weights, means, and diagonal covariances of GMM. $K$ and
    $N$ are the mixture number and the number of local features, respectively. $\mathbf{X}$
    denotes spatial-temporal local features in action videos.
    Perronnin \etal \cite{Fisher10} proposed an improved Fisher vector as follows,

    \begin{equation}
     v_{\mu,k} = \frac{1}{N\sqrt{\pi_k}}\sum_{i=1}^N{q_i(k)\Bigg( \frac{\mathbf{x_i} - \mu_k}{\sigma_k} \Bigg)}
    \end{equation} 
    \begin{equation} 
     v_{\sigma,k} = \frac{1}{N\sqrt{2 \pi_k}}\sum_{i=1}^N{q_i(k)\Bigg( \frac{(\mathbf{x_i} - \mu_k)^2}{\sigma_k} - 1 \Bigg)}
    \end{equation}

    where $q_i(k)$ is the posterior probability associating $\mathbf{x}_i$ to the $k$ Gaussian and is given by,
    \begin{equation}
     q_i(k) = \frac{\pi_k \mathcal{N}(\mathbf{x}_i;\mu_k,\Sigma_k)}{\sum_{n=1}^{K}{\pi_n \mathcal{N}(\mathbf{x}_i;\mu_n,\Sigma_n)}}
    \end{equation}
    
    The final Fisher vector is the concatenation of all $ v_{\mu,k}$ and $ v_{\sigma,k}$ 
    and is of $2Kd$ dimension. 
    Power normalization followed by $l_2$ normalization is applied to the FV and it 
    gives the best performance on image classification \cite{chatfield11devil} and 
    video-based action recognition \cite{Wang13iccv}.

    \begin{figure*}[t]
      \begin{center}
	\includegraphics[width=1.95\columnwidth,height=1\columnwidth]{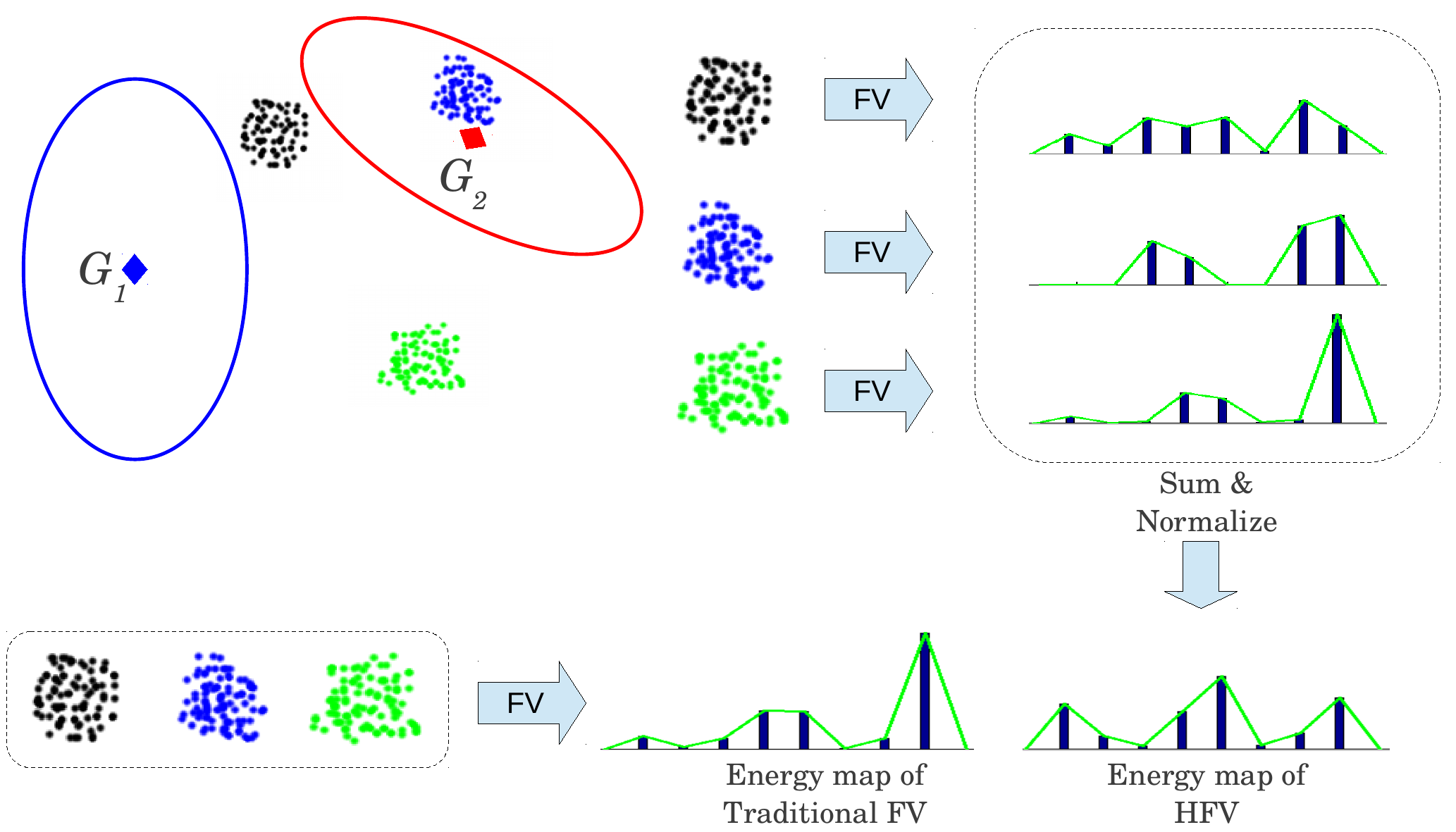}
      \end{center}
      \caption{Illustration of the difference in the energy distribution in Hyper Fisher Vector and traditional Fisher Vector 
               via a toy example (see section~\ref{sec:HFVbetter} for details).
               The HFV retains the individual energy maps of LFVs better and hence is more loyal to the contribution of 
               different feature clusters. In comparison, the traditional FV is biased towards the
               contribution of some features in the set.}
      \label{fig:hfv}
    \end{figure*}

    \subsection{Constructing Hyper-Fisher Vector}
    \label{sec:hfv_approach}
     The FV encoding results in high-dimensional feature vectors with less number of 
     Gaussians in the mixture and thus yields
     performance improvement when linear classifiers are used. However, the traditional FV 
     encoding aggregates the local features of
     an action video by sum pooling over the entire video. Such a representation cannot directly 
     represent higher complex structures. One way to alleviate this shortcoming is to use local 
     pooling and then pool the intermediate FVs. This global-local approach improves the performance
     of the FV encoding.

     The framework of our approach in comparison to the traditional FV encoding
     is illustrated in figure~\ref{fig:proposed_hfv}. Let $\mathbf{X} = (\mathbf{x}_1, 
     \mathbf{x}_2,  \ldots, \mathbf{x}_N)  \in \mathbb R^{d\times N}$
     be the local features (\eg HOF, HOG, MBH) obtained from the video of size $W\times H\times L$. 
     Then we compute the $k$-means cluster memberships for each feature $\mathbf{x}_i$ using a pre-learned
     dictionary codebook of size $K_1$ from the training set. 
     Let $C = (c_1, c_2, \ldots , c_N)$ be the cluster memberships of the
     features $\mathbf{X}$. Let there be $k_c$ clusters with non-zero members out of $K_1$ clusters.
     For each non-zero cluster, local Fisher Vectors, denoted by $LFV_i$ ($i=1 \ldots k_c$), are computed
     using a pre-learned GMM with mixture size $K_2$ in training set. The local Fisher Vectors are summed
     to result in the Hyper-Fisher Vector (denoted by $HFV$) representation of the video. The $HFV$ is power normalized 
     and $l_2$ normalized as in the case of traditional FV. The LFVs and the HFV are of length $2K_2d$.
      
     Algorithm~\ref{alg:hfv} gives the pseudocode for computing the HFV descriptors for a video. 
     $\mu$, {} $\Sigma$, {} $\pi$ in the psuedocode represent the mean, diagonal covariances 
     and mixing probabilities of the Gaussians in the pre-learned mixture. $FisherVectorCompute$
     computes the FV representation of the input features using the GMM parameters.

   \begin{algorithm}
    \caption{Compute Hyper-Fisher vector descriptor}
    \begin{algorithmic}
       \STATE {\bf Input:} {$\{\mathbf{x}_i\}_{i=1}^{N}$, {} $\{c_i\}_{i=1}^{N}$, {} $K_1$, {} $\mu$, {} $\Sigma$, {} $\pi$ }
       \STATE
       \STATE {\bf Output:} $HFV$ 
       \STATE
       \STATE Initialize $HFV$ to $\mathbf{0}$
       \FOR{ $k = 1$ to $K_1$}
          \STATE $F = \{\mathbf{x}_i$  $ | $  $ c_i = k\}$
          \IF {$F = \phi $}
              \STATE {\bf continue}
          \ENDIF
          \STATE $LFV = $ $FisherVectorCompute(F,\mu, \Sigma,\pi)$
          \STATE $HFV = $ $HFV + LFV$ 
       \ENDFOR 
       \STATE $Power$ {} $normalize$ {} $HFV$
       \STATE $l_2$ {} $normalize$ {} $HFV$
      
    \end{algorithmic}
    \label{alg:hfv}
   \end{algorithm}

   \begin{figure*}[t]
      \begin{center}
	\includegraphics[width=1.99\columnwidth,height=0.55\columnwidth]{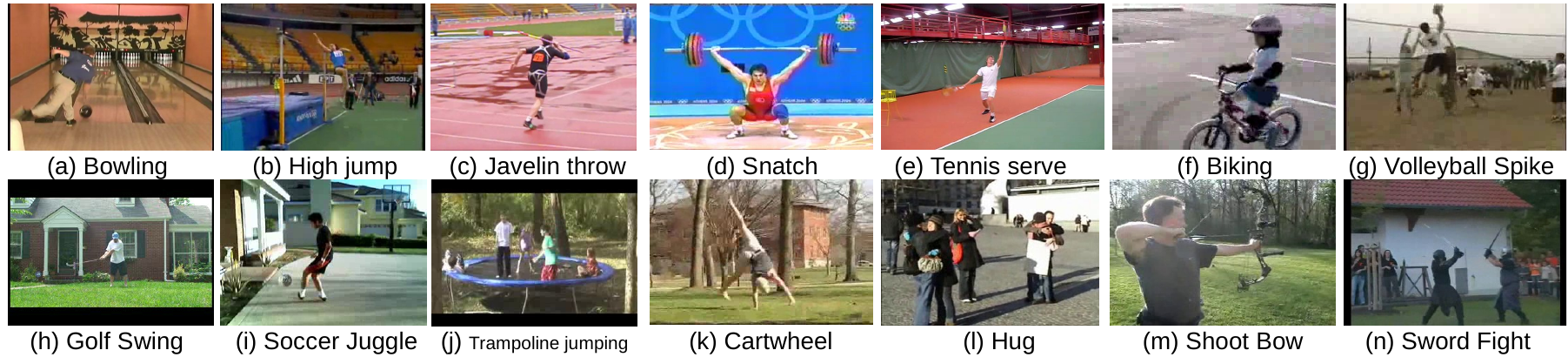}
      \end{center}
      \caption{Samples from the datasets. (a)-(e) is from Olympic Sports, (f)-(j) is from Youtube (and UCF$50$) and 
               (k)-(n) is from HMDB$51$ datasets.}
      \label{fig:dataset}
    \end{figure*} 
    
   \subsection{Why Hyper-Fisher Vectors are superior?}
   \label{sec:HFVbetter}
   In the last subsection, we showed the approach to construct Hyper-Fisher Vectors. 
   In this subsection, we analyze why 
   the HFVs are superior in comparison to the traditional FVs. We illustrate the difference between the 
   two using a toy example (figure~\ref{fig:hfv}).
   We consider $2$ Gaussians in the mixture and three clusters of features in the feature set.
   The Gaussians are centered at $(0,0)$ and $(4,4)$ with diagonal variances $(0.5,4)$ and $(0.5,1)$ along 
   $(x,y)$ directions respectively.
   The three clusters are chosen such that one cluster (centered at $(4,4.5)$ and shown in blue) 
   is well with in one of the Gaussians, 
   the second cluster of features (centered at $(2,2.5)$ and shown in black)
   is in between
   the two Gaussians and the third cluster (centered at $(3.5,-0.5)$ and shown in green) is slightly far away 
   from both the Gaussians. All the features are pooled together and the traditional FV
   representation is obtained. Since we consider $2$ Gaussians in 2$d$ space, the length of the FV is $8$.
   Standard representation of FV is used where the mean deviation components form the first-half of the FV 
   followed by the variance deviation components.

   The energy distribution for the traditional FV among the 
   mean and covariance deviation components is shown in figure~\ref{fig:hfv}. 
   For the HFV representation, the clusters are represented by three different LFVs 
   and summed and normalized to obtain the HFV. The energy distributions for each 
   LFV and the HFV are also shown in the figure. The black cluster of features has even 
   distribution of energy among its LFV components and across Gaussians since it is near to both of them.
   The blue cluster is with in the second Gaussian and hence only those components corresponding to second 
   Gaussian in the associated LFV are high. The green cluster is slightly far from both the Gaussians and 
   has higher energy in the covariance deviation components as compared to the mean deviation 
   components in its LFV. It can be seen that the energy in the covariance deviation components 
   is higher than the mean deviation components in the traditional FV. Whereas, in comparison, the HFV has more
   energy in its mean deviation components than their counterparts in traditional FV. The energy distribution
   in HFV is more loyal to the individual distributions in LFVs and hence to the feature clusters.
   Hence, the HFV represents the feature set better than the traditional FV.
   
   The similarity score (using the linear kernel) between the HFV and the FV shown in the figure is around $0.8$.
   This depends on the range/width of the 
   clusters. Wider the clusters, higher is the similarity between HFV and FV. Quantitative results on the energy
   distribution and the similarity between HFV and FV are given in the experimental 
   results section (section ~\ref{sec:exp_results}).

   \section{Experimental Setup}
   \label{sec:exp_setup}
   In this section, the details of the experimental setup with various parameter settings are provided.
   The datasets used for evaluating the approach are presented in section~\ref{sec:datasets}.

   In the following experiments, improved trajectories and associated descriptors are extracted 
   using the code from Wang \cite{Wang13iccv}. Default parameters are used to extract the trajectories.
   For the $k$-means clustering (required for HFV encoding), the size of the 
   codebook is chosen to be $4000$ and is learnt using randomly sampled $100,000$ features. 
   For the traditional FV and HFV encodings, the dimensionality of these descriptors 
   is reduced by half using PCA. 
   For the traditional FV, a GMM of size  $256$ is 
   learnt using randomly sampled $100,000$ features. The same GMM is used for HFV encoding as well.
    A linear SVM is used for classification. 
    We use a \emph{one-vs-all} approach while training the multi-class classifier.
    
    The baseline for our Hyper-Fisher Vector encoding is the traditional Fisher Vector encoding.
    We also experiment with different power normalizations for the traditional FV encoding and compare 
    against the proposed encoding.
    
    
    \subsection{Datasets}
    \label{sec:datasets}
    We perform the experiments on four action recognition datasets and report the results. 
    The datasets used for evaluating our work are Olympics Sports, UCF$11$ (also called Youtube dataset),
    UCF$50$ and HMDB$51$.
    Few samples from the datasets are shown in figure~\ref{fig:dataset}.

    The {\bf Olympic Sports} dataset \cite{Niebles10Olympics} contains videos of athletes 
    practicing different sports collected from
    Youtube. It contains $16$ sports action categories and over $750$ videos. 
    Some of the classes are \emph{bowling}, \emph{high jump},  
    \emph{shot put}, \emph{tennis serve}. We use the test-train splits provided by the authors for
    evaluation and report the mAP over all the classes.

    \begin{table*}[t]
      \begin{center}
      \begin{tabular}{|l|c|c|c|c|}
      \hline
      Method & Olympic Sports & Youtube & UCF$50$ & HMDB$51$ \\
      \hline\hline
      FV           &  91.1\%   & 90.7\%   &  91.2\%  & 57.2\%  \\
      FV ($p < 0.5$)   &  91.7\%   & 91.8\%   & 92.1\%   & 58.5\%  \\
      \hline
      {\bf Hyper FV}    &  92.8\%   & 92.9\%   &  93.0\%  & 60.1\%   \\
      \hline
      \end{tabular}
      \end{center}
      \caption{Performance comparison on the three datasets using baseline FV
               and the proposed Hyper-FV encodings. $p < 0.5$ indicates stronger power normalization 
               used for encoding.}
      \label{tab_hfv_baseline}
    \end{table*}
    
    The {\bf Youtube dataset} \cite{JLiu09} is collected from YouTube videos. It contains
    $11$ action categories. Some of the actions are \emph{basketball shooting},
    \emph{riding horse}, \emph{cycling}, \emph{walking (with a dog)}. 
    A total of $1,168$ video clips are available.
    As in \cite{JLiu09}, we use Leave-One-Group-Out cross-validation and report the
    average accuracy over all classes.
    
    The {\bf UCF$50$ dataset} \cite{Reddy12} is an extension of the Youtube dataset and contains
    a total of $6618$ clips from $50$ action categories. We apply the 
    Leave-One-Group-Out cross-validation ($25$ cross-validations) as suggested by the 
    authors \cite{Reddy12} and report the average accuracy over all classes.

    The {\bf HMDB51} action dataset  \cite{Kuehne11hmdb} 
    is collected from various sources, mostly from movies, and from public databases such as 
    YouTube and Google videos. 
    The dataset contains $6766$ clips categorized into $51$ action classes, each containing a minimum of $101$ clips. 
    The action categories can be grouped into general facial actions, general body movements with and without object 
    interactions and human interactions. We use the original
    $3$ train-test splits provided by the authors for evaluation. Each split contains $70$ videos 
    and $30$ videos from every class for training for testing respectively. The average classification accuracy 
    over the three splits is reported.

    \section{Experimental Results}
    \label{sec:exp_results}
    We conduct different experiments over the datasets to evaluate the performance of the proposed encoding.
    The results of the experiments Hyper-FV encoding are tabulated in Table~\ref{tab_hfv_baseline}.
    The traditional FV encoding is the baseline for the Hyper-FV encoding.
    Since, the HFV encoding involves two power normalizations, we also compare against
    the traditional FV encoding with stronger power normalizations ($p<0.5$).
    
\begin{figure}[t]
      \begin{center}
	\includegraphics[width=0.99\columnwidth,height=0.85\columnwidth]{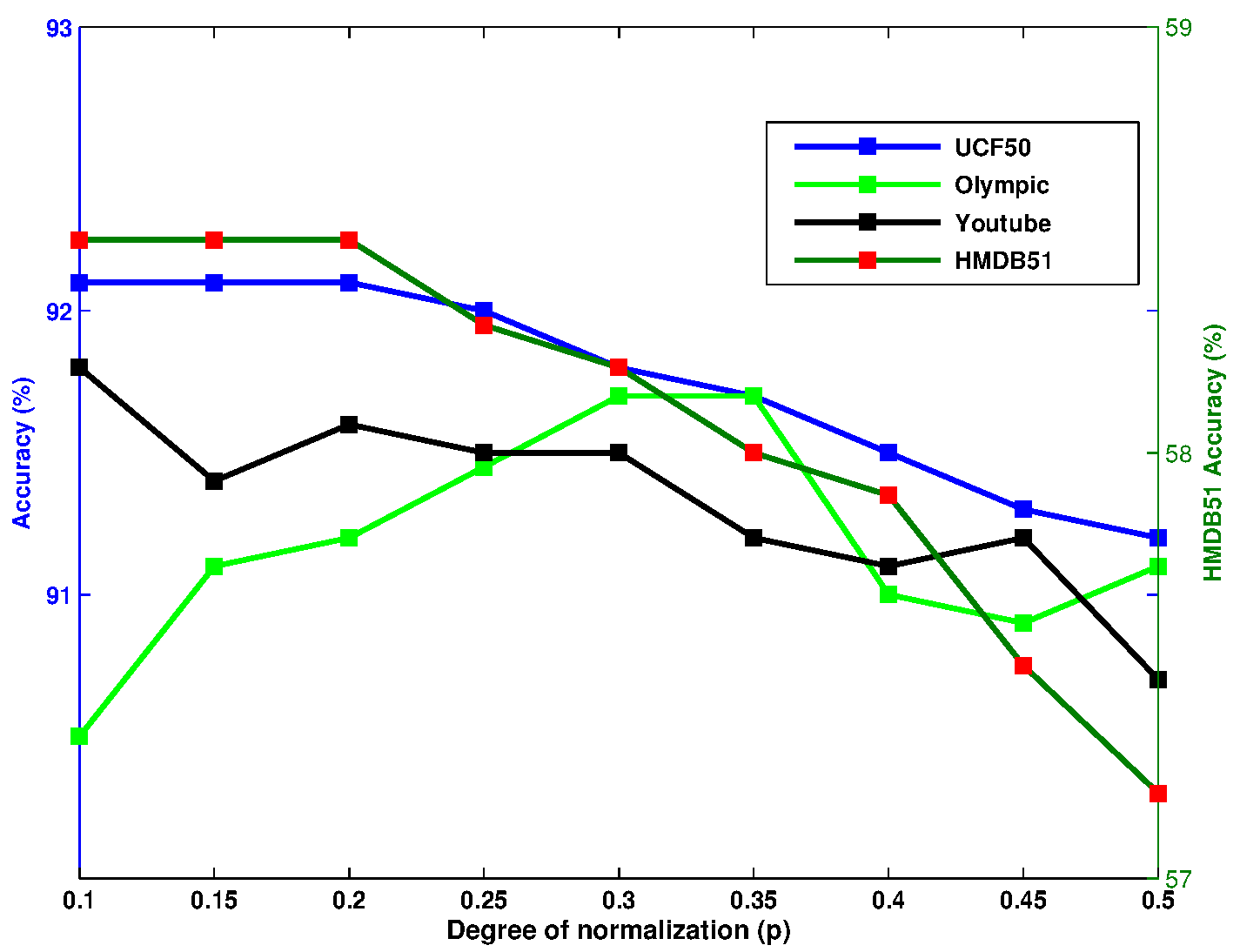}
      \end{center}
      \caption{ Plots showing the variation of the performance of FV encoding as the degree of power normalization
               is varied (HMDB$51$ accuracy on the right $y$-axis).}
      \label{fig:norm_vary}
    \end{figure}

    We observe from table~\ref{tab_hfv_baseline} that Hyper-FV performs better compared to 
    the traditional FV encoding on all the datasets.   
    The improvement is around $2\%$ for the Olympic Sports, Youtube and UCF$50$ datasets
    and $3\%$ for HMDB$51$ dataset. 
    The performance of the FV encoding also improves when a stronger power normalization
    is used. The table~\ref{tab_hfv_baseline} reports the best performance for each dataset when $p<0.5$. 
    Figure~\ref{fig:norm_vary} shows the variation in the performance of FV encoding as the normalization power
    is varied. Except for Olympic Sports dataset, the accuracy improves as we decrease $p$ from $0.5$ to $0.1$ and 
    the best performance is achieved when $p$ is in the range $0.1$ to $0.2$. For the Olympic Sports dataset,
    the maximum is reached for $p=0.35$ below which the accuracy decreases. 
    Even though there is an improvement in the performance when a stronger power normalization is used,
    the performance of the HFV encoding is still better, in general, by $1-1.5\%$ for the four datasets
    as noted from table~\ref{tab_hfv_baseline}.
    This shows that a simple modification in the way the Fisher Vectors are encoded 
    can improve the performance on challenging datasets like UCF$50$ and HMDB$51$. 
    
    \begin{figure*}[t]
      \begin{center}
	\includegraphics[width=1.99\columnwidth,height=0.55\columnwidth]{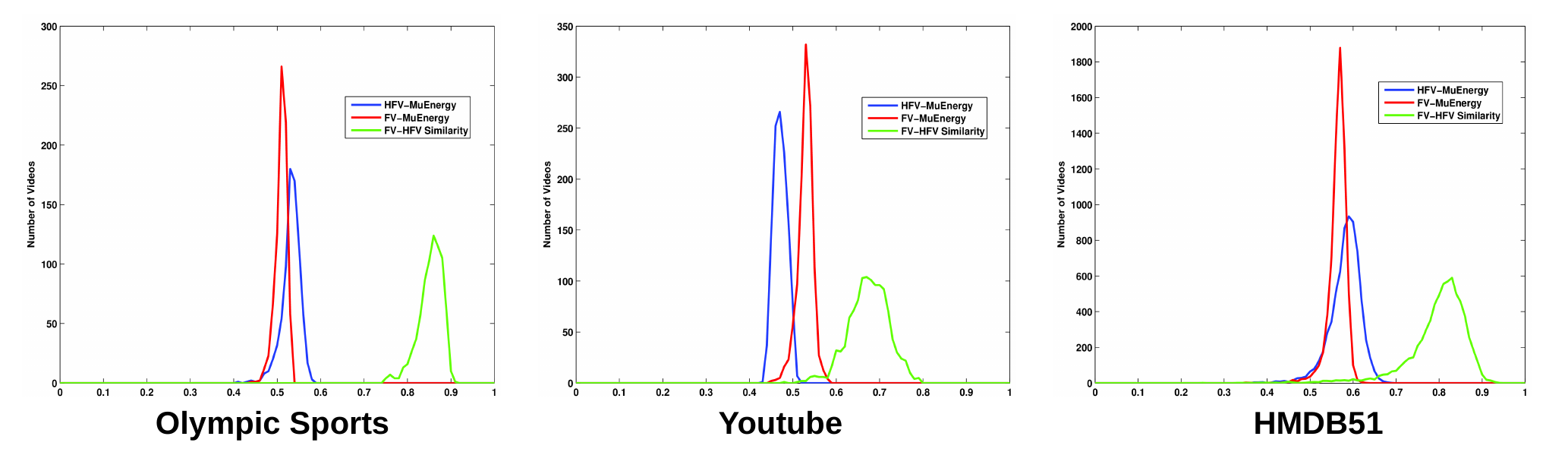}
      \end{center}
      \caption{Plots showing the proportion of the unit energy in mean deviation components of FV and HFV encoding is shown in red and blue respectively.
               The green curve shows the similarity measure between FV and corresponding HFV of videos in the datasets. The $y$-axis is the number
               of videos in the datasets (best viewed when zoomed).}
      \label{fig:energy}
    \end{figure*} 
    
    Figure~\ref{fig:energy} shows three plots. Each plot has three functions plotted. The red curve
    depicts the number of videos in the dataset having different energy proportions in the
    mean deviation components of the Fisher vector 
    representation of the video. The blue curve depicts the same for the Hyper-Fisher vector representation. Since the 
    total energy in the Fisher vectors sums to $1$, the remaining energy is present in the covariance deviation components
    of the respective representations. We can observe that the HFV representation in general has 
    mean deviation components with 
    broader energy range than the corresponding FV representations of the videos. 
    The FV red curves are more sharper than the 
    HFV blue curves for each dataset. This shows that the HFV representation has better variations 
    in its components and
    represents the video actions better. The third curve (in green) shows the 
    similarity scores range for the videos in the 
    dataset. The similarity scores are between corresponding FV and HFV of the videos. 
    The green curve indicates that more than 
    $50\%$ of the videos in the HMDB$51$ dataset have their FV-HFV similarity less than $0.85$. 
    For the Youtube dataset, the similarity scores are centered around $0.7$.
    This indicates the difference in the 
    representations.

%
      
    \subsection{Robustness of HFV encoding}
\begin{figure}[h]
      \begin{center}
	\includegraphics[width=0.99\columnwidth,height=0.85\columnwidth]{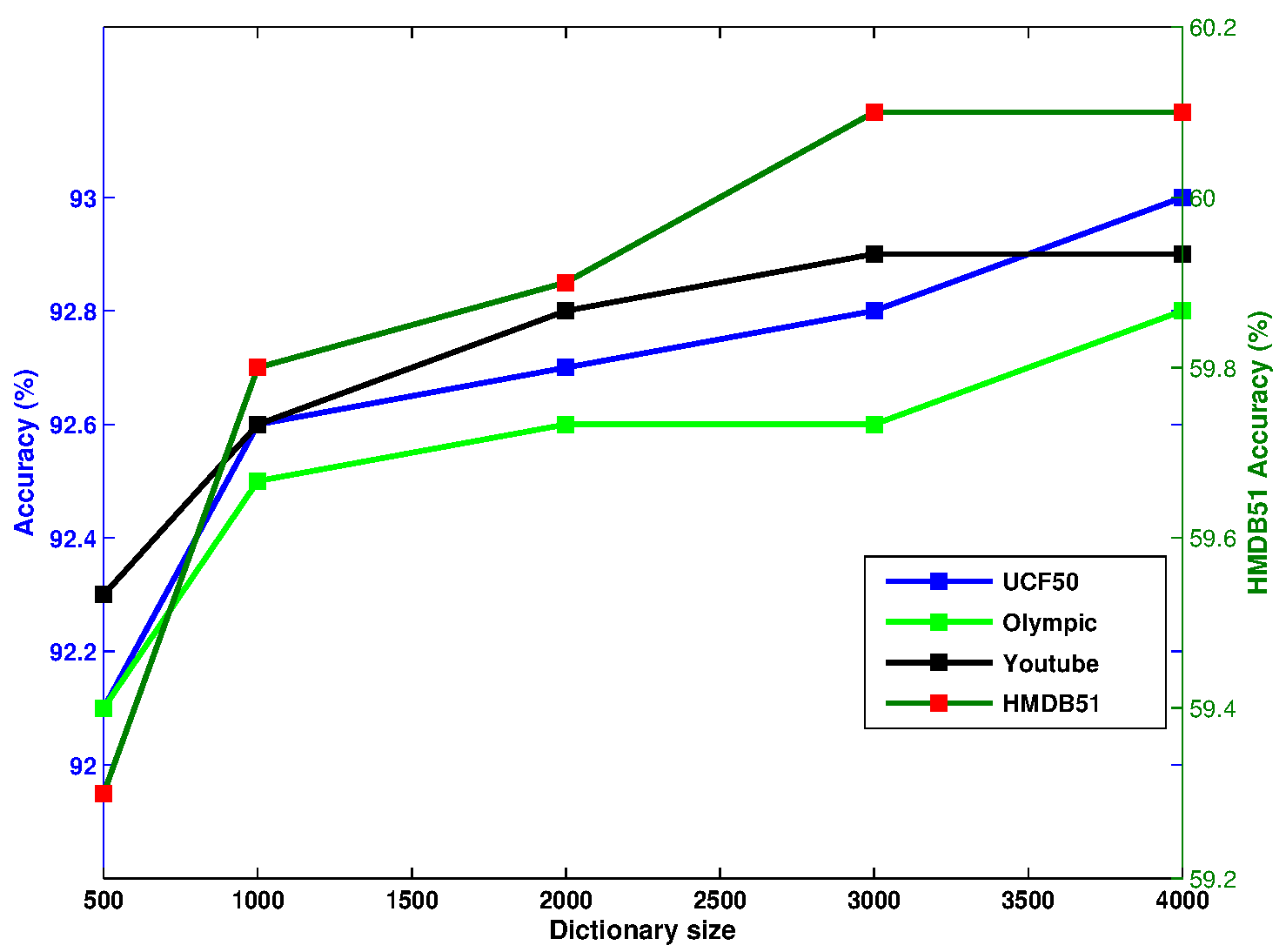}
      \end{center}
      \caption{ Plots showing the variation of the performance of Hyper-FV encoding as the codebook size is 
               varied (HMDB$51$ accuracy on the right $y$-axis).}
      \label{fig:hfv_robust}
    \end{figure}

    We conduct experiments to test the robustness of the proposed Hyper-FV encoding. The dictionary size of the 
    $k$-means clustering is varied and the performance of the HFV on the datasets is plotted. Figure~\ref{fig:hfv_robust}
    shows the variation of performance of the HFV encoding as a function of the dictionary size. 
    The dictionary size is varied from $500$ to $4000$. We can see that the accuracy variation is marginal 
    (within $1$ percent)
    and the HFV encoding performs well even with lower codebook sizes. This shows that the HFV encoding is 
    robust to the codebook size. 
    
    \subsection{Comparison with approaches in literature}
    We compare the results of our method with the recent results reported in literature for each dataset.
    It is tabulated in table~\ref{tab_all_res}. 
    For the purpose of a fair comparison, approaches involving deep networks for action recognition are not 
    compared here. 
    The improvements for Olympic Sports, Youtube and UCF$50$ datasets are around  $2\%$ and our method improved the performance on
    the more challenging HMDB$51$ dataset by $3\%$ in comparison to the other approaches. In particular, 
    Wang \etal \cite{Wang13iccv} use the Fisher Vector encoding and in comparison, the proposed encoding performs better.
    This shows that our HFV encoding can be used to substitute the original FV encoding for improved performance in 
    various applications.

    \begin{table*}[t]
      \begin{center}
      \begin{tabular}{|lc|lc|lc|lc|}
      \hline
      Olympic Sports & & Youtube & & UCF$50$ & & HMDB$51$ & \\
      \hline\hline
         Gaidon \etal \cite{gaidon12recognizing} & 82.7\% & Wang \etal \cite{Wang13ijcv} & 85.4\% & Wang \etal \cite{Wang13ijcv} & 84.5\% & Wang \etal \cite{Wang13ijcv} & 46.6\% \\
         Jain \etal \cite{Jain13cvpr} & 83.2\% & Liu \etal \cite{JLiu09} & 71.2\% &  Shi \etal \cite{Shi13cvpr} & 83.3\% & Zhu \etal & 54.0\% \\
         iDT+FV \cite{Wang13iccv} &  91.1\% & iDT+FV \cite{Wang13iccv} & 90.7\% & iDT+FV \cite{Wang13iccv} & 91.2\% & iDT+FV \cite{Wang13iccv} & 57.2\%\\  
      \hline
      {\bf Proposed}   & 92.8\%  &  {\bf Proposed} & 92.9\% & {\bf Proposed} & 93.0\% & {\bf Proposed} & 60.1\%\\
      \hline
      \end{tabular}
      \end{center}
      \caption{Comparison of our proposed approach with other approaches on Olympic Sports,
               Youtube, UCF$50$ and HMDB$51$ datasets.}
      \label{tab_all_res}
    \end{table*}

    \section{Conclusion}
    \label{sec:conclusion}
    In conclusion, we have developed a novel Hyper-Fisher Vector encoding which embeds the Bag-of-Words encoding
    into the Fisher Vector encoding. In this work, the Hyper-FV encoding has been used to represent actions
    in videos. We evaluated our approaches on challenging datasets such as UCF$50$ and HMDB$51$ 
    and the Hyper-FV encoding was shown to perform better than the FV encoding. Thus the proposed encoding 
    can be used in place of the FV encoding in different applications for better representation and can
    also be used in deep networks, such as deep Fisher networks for action recognition.

    
%

{\small
\bibliographystyle{ieee}
\bibliography{refer}
}

\end{document}